\documentclass[conference]{IEEEtran}
\usepackage[top=0.75in, bottom=1in, left=0.625in, right=0.625in]{geometry}
\IEEEoverridecommandlockouts
\usepackage{amsmath}
\usepackage{soul}
\usepackage[section]{placeins}

\usepackage{verbatim} 
\usepackage{pifont}
\usepackage{times,amsmath,color,amssymb,graphicx, epstopdf, epsfig,cite,psfrag,subfigure,algorithm}
\usepackage[noend]{algpseudocode}
\usepackage{enumerate}
\usepackage{xcolor}
\usepackage{lipsum} 
\usepackage{setspace} 

\usepackage{color}
\begin{document}

	
	%
	\title{


Accelerating Privacy-Preserving Federated Learning in Large-Scale LEO Satellite Systems


}
	%
	%
	%
	


\author{\IEEEauthorblockN{Binquan Guo$^{a,b,c}$, Junteng Cao$^{d}$, Marie Siew$^{c}$, Binbin Chen$^{c}$, Tony Q. S. Quek$^{c}$, and Zhu Han$^{e}$ }\\
\IEEEauthorblockA{$^a$School of Telecommunication Engineering, Xidian University, Xi'an 71071, P. R. China\\
$^b$Tianjin Artificial Intelligence Innovation Center (TAIIC), Tianjin 300457, P. R. China\\
$^c$Pillar of ISTD, Singapore University of Technology and Design, 487372, Singapore\\
$^d$School of Electrical and Electronic Engineering, Nanyang Technological University, Singapore \\
$^e$Department of Electrical and Computer Engineering, University of Houston, Houston, TX 77004 USA
}
}

	\maketitle


	
	\begin{abstract}
		

Large-scale low-Earth-orbit (LEO) satellite systems are increasingly valued for their ability to enable rapid and wide-area data exchange, thereby facilitating the collaborative training of artificial intelligence (AI) models across geographically distributed regions. Due to privacy concerns and regulatory constraints, raw data collected at remote clients cannot be centrally aggregated, posing a major obstacle to traditional AI training methods. Federated learning offers a privacy-preserving alternative by training local models on distributed devices and exchanging only model parameters. However, the dynamic topology and limited bandwidth of satellite systems will hinder timely parameter aggregation and distribution, resulting in prolonged training times. To address this challenge, we investigate the problem of scheduling federated learning over satellite networks and identify key bottlenecks that impact the overall duration of each training round. We propose a discrete temporal graph–based on-demand scheduling framework that dynamically allocates communication resources to accelerate federated learning. Simulation results demonstrate that the proposed approach achieves significant performance gains over traditional statistical multiplexing-based model exchange strategies, reducing overall round times by 14.20\% to 41.48\%. Moreover, the acceleration effect becomes more pronounced for larger models and higher numbers of clients, highlighting the scalability of the proposed approach.

\end{abstract}


\vspace{-3 mm}

\section{Introduction}
%
%
%
%

Low-Earth-orbit (LEO) satellite networks are emerging as a key infrastructure for enabling seamless connectivity and knowledge sharing across the globe \cite{al2023artificial,le2024survey,chen2022satellite}. Unlike traditional geostationary satellites, LEO satellites operate at altitudes ranging from 500 to 2,000 kilometers, significantly reducing communication latency and enabling closer integration with terrestrial networks \cite{hassan2020dense}. In recent years, several commercial and governmental initiatives have launched mega-constellations comprising thousands of LEO satellites \cite{cao2022edge, hu2024performance,guo2025enabling}. For instance, by the end of July 2025, SpaceX's Starlink has deployed over 7,000 satellites in orbit, targeting global broadband coverage. These LEO satellite constellations are designed not only to deliver internet services, but also to serve as an infrastructure layer for supporting emerging intelligent applications on a global scale.
The capabilities of modern LEO constellations are further enhanced by the incorporation of inter-satellite links, which allow satellites to relay data among themselves without relying solely on ground stations \cite{lai2022spacertc}. The inter-satellite communication improves routing flexibility, reduces ground relay latency, and supports persistent end-to-end communication. At the user end, direct-to-device technologies are gaining momentum, with satellite-to-smartphone communication becoming commercially viable, as demonstrated by recent partnerships between satellite operators and mobile manufacturers \cite{le2024survey}. These trends collectively position LEO satellite networks as a promising platform for delay-sensitive and AI-native applications, particularly in regions lacking terrestrial infrastructure \cite{lin20215g}.

While LEO satellite systems have the capability to enable collaborative machine learning across regions, privacy concerns, regulatory requirements, and data protection policies such as the General Data Protection Regulation (GDPR) and the EU Artificial Intelligence Act (EU AI Act), make centralized training challenging\cite{du2025hyshield}. In many domains, such as healthcare, finance, and critical infrastructure, the raw data generated at distributed sites cannot be transmitted to centralized servers for training due to legal, ethical, or operational limitations. 
These constraints hinder collaborative learning using data generated across regions, especially for remote or underserved areas where terrestrial connectivity is inadequate and satellite networks become indispensable for supporting federated learning ~\cite{zhang2022progress}.
Federated learning offers a promising solution to this problem \cite{chen2022satellite}. Instead of transferring raw data, each participating device or data source performs local training and shares only local model parameters with a coordinating server \cite{du2024rdp}. This approach ensures that sensitive data remains at its point of origin while still contributing to the construction of a globally shared model~\cite{razmi2024board, zhai2023fedleo}. The decentralized nature of federated learning aligns naturally with the architecture of LEO satellite networks~\cite{shi2024satellite}. By transmitting only model updates rather than entire datasets, federated learning significantly reduces the communication overhead and conserves limited satellite bandwidth. As a result, the combination of federated learning with satellite-based infrastructures creates a scalable and privacy-preserving framework for collaborative model training, thereby enabling AI models to be trained and updated across geographically dispersed environments without violating data confidentiality.

Despite the promising synergy between federated learning and satellite networks, several unique challenges arise that do not typically appear in terrestrial networks \cite{razmi2024board}. One primary issue is the \textbf{inherently limited bandwidth} available on satellite communication links, especially in satellite-direct-to-device scenarios. 
While inter-satellite laser links are under active development and can exceed 1 Gbps in tests, end-to-end performance remains bottlenecked by the satellite-to-device segment, where recent Starlink tests report downlink speeds of only around 17 Mbps (according to Starlink).
Unlike ground networks where bandwidth can be more abundant, satellite links must operate \textbf{within strict capacity constraints} \cite{guo2024network}. 
Furthermore, broadcasting model updates to \textbf{geographically distant and widely dispersed} clients requires maintaining dedicated end-to-end communication channels over dynamic satellite networks \cite{zhang2023time}. 
This limitation leads to contention among multiple model transmissions competing for scarce bandwidth resources, thus resulting in slower model dissemination. The prolonged transmission times directly affect the federated learning process. In particular, local training cannot begin until the updated global model is fully received, and uploading trained model parameters back to the server is further delayed. Consequently, each training iteration in satellite-based federated learning consumes significantly more time than in terrestrial settings. To address these challenges, it is essential to design \textit{specialized scheduling schemes that efficiently allocate communication resources, minimize transmission delays, and thereby accelerate the overall process} of federated learning in satellite networks.

\begin{figure}[!htbp]
	\centering
 	\includegraphics[width=80mm]{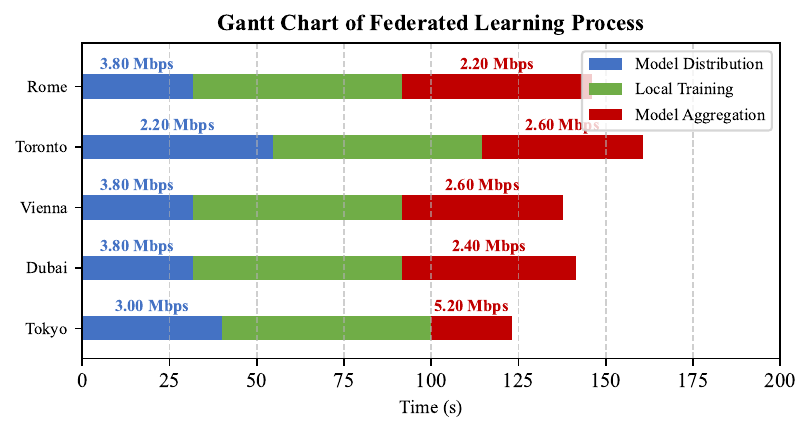}\\
		\vspace{-3 mm}
\caption{Gantt chart for the timeline of a federated learning round using the traditional statistical multiplexing scheduling scheme, where model transmission durations are prolonged due to shared bandwidth among multiple clients.}
\label{fig:old_method}
\end{figure}

\begin{figure}[!htbp]
	\centering
 	\includegraphics[width=80mm]{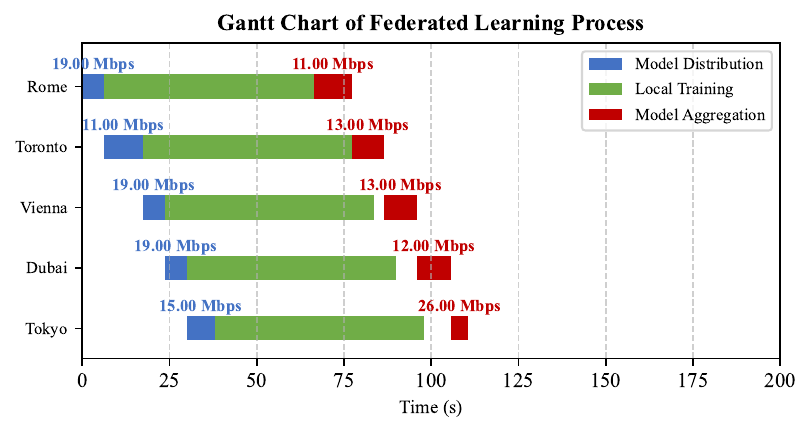}\\
    		\vspace{-3 mm}
\caption{Gantt chart for the timeline of a federated learning round using the proposed on-demand scheduling scheme, where model exchange is optimized by scheduling model parameter transmissions in an ordered manner.}
\label{fig:new_method}
\end{figure}

In fact, the primary innovation of our work lies in leveraging the temporal graph-based on-demand scheduling model to enhance the efficiency of model transmission for federated learning in satellite networks, thereby outperforming traditional statistical multiplexing approaches. The distinction between these two strategies and the resulting performance gains of our method are illustrated in Figs.~\ref{fig:old_method} and \ref{fig:new_method}. A comparison of the timelines in these figures shows that the traditional statistical multiplexing approach requires over 150 seconds to complete one training round, whereas our on-demand scheduling scheme accomplishes the same round in under 125 seconds, demonstrating a substantial reduction in round makespan. Given that practical federated learning applications require numerous training rounds for convergence, this per-round time saving accumulates into a significant reduction in total training duration, enabling faster model convergence and more efficient utilization of the satellite network's limited resources.

In this article, we investigate the problem of scheduling federated learning tasks over LEO satellite networks to overcome the transmission bottlenecks and accelerate the training process. We begin by analyzing the key factors that contribute to prolonged iteration times, focusing on the \textit{interplay between limited satellite bandwidth  and the contention among multiple model parameter transmissions}. To address these issues, we propose a discrete temporal graph–based on-demand scheduling framework that dynamically allocates communication resources to optimize the timing of model parameter exchanges. Our approach adapts to the time-varying topology and capacity fluctuations inherent in satellite networks. Simulation results demonstrate that the proposed scheduling method significantly outperforms traditional statistical multiplexing approaches, achieving substantial reductions in iteration makespan. Moreover, the benefits of our method increase with the size of the model being trained, highlighting its scalability and practical applicability for large-scale federated learning in satellite environments.
The main contributions of this article are summarized as follows:

\begin{itemize}
\item We identify and analyze the critical bottlenecks in federated learning over LEO satellite networks, emphasizing the impact of limited bandwidth and model transfer contention on model parameter transmission delays.

\item We propose a discrete temporal graph–based on-demand scheduling framework that dynamically allocates satellite communication resources to optimize the timing of model parameter exchanges and accelerate the overall process.

\item Our scheduling approach adapts to the dynamic and time-varying topology of satellite networks, effectively reducing iteration makespan and improving training efficiency.

\item Simulation results show significant performance improvements over traditional statistical multiplexing, with overall round times reduced by 14.20\% to 41.48\%, and even greater gains observed for larger models and higher numbers of clients, validating the scalability and effectiveness of the proposed method.

\end{itemize}

The remainder of this article is organized as follows. Section \ref{sec:system} introduces the system model and analyzes the challenges inherent in federated learning over LEO satellite networks. Section \ref{sec:method} presents the proposed temporal graph–based scheduling framework and details its design and workflow. Section \ref{sec:evaluation} evaluates the performance of the proposed method through extensive simulations. Finally, Section \ref{sec:conclusion} concludes the article.





\section{System Model and Problem Statement} \label{sec:system}

\begin{figure}[t!]
	\centering
 	\includegraphics[width=79mm]{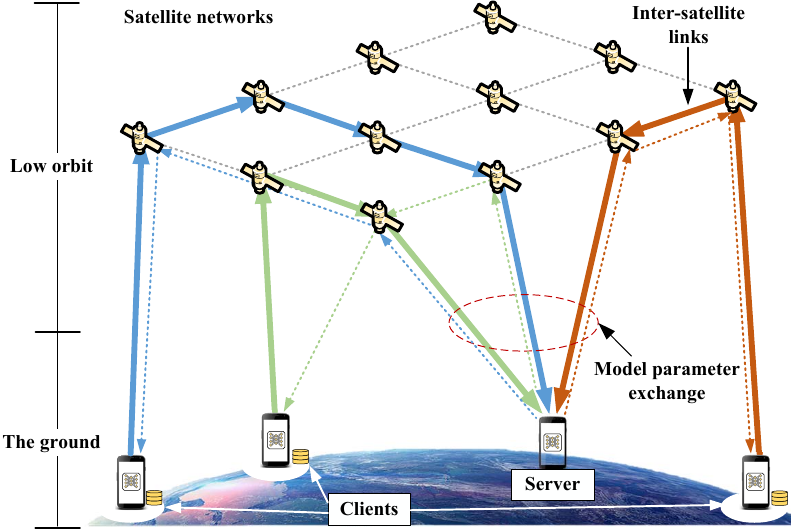}\\
    		\vspace{-3 mm}
	\caption{The scenario of federated learning in LEO satellite networks. }
\label{fig:satelliteFL_networks}
\end{figure}


\subsection{Satellite-Based Federated Learning System Model}

We consider a satellite-based federated learning system composed of an LEO satellite constellation and a set of user devices distributed globally, as illustrated in Fig.~\ref{fig:satelliteFL_networks}. The satellite network consists of a set of LEO satellites denoted by $\mathcal{O} = \{\mathbb{O}_1, \mathbb{O}_2, ..., \mathbb{O}_{|\mathcal{O}|}\}$ and a set of user devices denoted by $\mathcal{U} = \{\mathbb{U}_1, \mathbb{U}_2, ..., \mathbb{U}_{|\mathcal{U}|}\}$. The satellites are dynamically interconnected via inter-satellite links  and establish satellite-to-ground links with user devices when they are within line-of-sight. Due to the fast and periodic movement of LEO satellites, both the network topology and the connectivity between satellites and user devices change over time.
In this work, we focus on LEO satellites as relays for federated learning, with remote user devices relying on multi-hop paths for model exchanges due to the absence of terrestrial infrastructure.

\subsection{Federated Learning Task Model}

\begin{figure}[!htbp]
	\centering
 	\includegraphics[width=60mm]{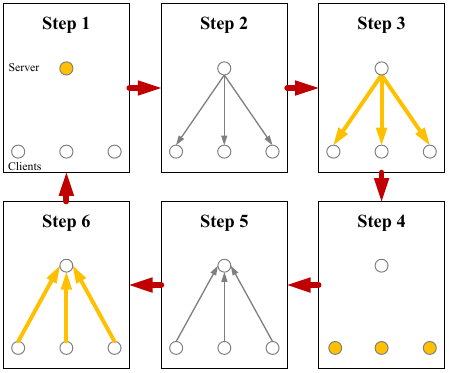}\\
    \caption{The six steps of federated learning in satellite networks, including 1) global model update, 2) distribution path computation, 3) model distribution, 4) local training, 5) aggregation path computation, and 6) model aggregation.}
\label{fig:fedSat_steps}
\end{figure}

In this work, we define a federated learning task as $F = \{\mathbb{S}, \mathcal{C}, \mathbb{M}\}$, where $\mathbb{S}$ denotes the server responsible for model distribution, updating, and aggregation. $\mathcal{C} = \{\mathbb{C}_1, \mathbb{C}_2, ..., \mathbb{C}_{|\mathcal{C}|}\}$ represents the set of local clients holding private datasets; and $\mathbb{M}$ denotes the AI model to be trained. In this study, each federated learning task involves training a single model. Based on the local dataset size and the computational capability of each client $\mathbb{C}_i$, the training duration is denoted by $\tau_{\mathbb{C}_i}$. Additionally, the data size of the model $\mathbb{M}$ required by client $\mathbb{C}_i$ is denoted by $d_{\mathbb{C}_i}$.
The entire federated learning process is carried out collaboratively between the server and the clients over multiple rounds, without the need to exchange raw data \cite{zhang2024srfl}. Instead, only model parameters are transmitted. Each global round consists of the following \textit{six steps}:

\textbf{Step 1: Global model update.}  
At the beginning of each global round, the server updates the global model using the aggregated results from the previous round or initializes a fresh model if it is the first round.

\textbf{Step 2: Routing path computation for global model distribution.}  
The server computes multi-hop routing paths to deliver the global model to each client via the satellite network. This involves determining feasible end-to-end paths from the server to the clients based on satellite topology, link availability, and propagation delays \cite{chen2024shortest}.

\textbf{Step 3: Global model distribution.}  
The server transmits the global model parameters to each client through the computed  paths. 
This step entails routing through multiple satellites to deliver the model individually to each client.

\textbf{Step 4: Local training.}  
Upon receiving the global model, each client performs local training using its private dataset. The end time of the training depends not only on the client’s computation power but also on the time it receives the model. Clients that receive the model earlier can start training sooner, potentially reducing the total round duration.

\textbf{Step 5: Routing path computation for model aggregation.}  
The server computes paths from each client to the server for uploading the locally updated models. This step also involves routing across multiple time slots, accounting for satellite mobility and dynamic link states.

\textbf{Step 6: Model aggregation.}  
Clients upload their locally trained model parameters to the server using the computed paths. Once all local models are collected, the server updates the global model. This process then repeats from \textbf{Step 1}.

As illustrated in Fig.~\ref{fig:fedSat_steps}, these six steps are executed repeatedly until the training process converges or meets the termination condition. Compared to terrestrial federated learning \cite{du2025dp}, satellite-based federated learning introduces new challenges, particularly in Steps 2 and 5, due to the dynamic topology and limited bandwidth of the satellite network. The entire process is coordinated by a ground control center, which communicates with satellites through a dedicated geostationary satellite link as in \cite{guo2023online}. The ground control center collects satellite network state information, computes routing paths for model exchange, and issues scheduling commands.


\subsection{Problem Statement}

The scheduling of federated learning in satellite networks introduces formidable challenges, primarily originating from the dynamic network topology and constrained communication resources. As previously established, these characteristics can lead to substantial latencies during the transmission of model parameters between the server and the distributed clients. Such latencies can become the dominant bottleneck in the training process. This extends the duration of each federated training round, thereby diminishing the overall efficiency and delaying the convergence of the global model.
The objective of this work is to overcome this critical bottleneck. 
By strategically coordinating the transmission of model parameters and the execution of local training tasks, our goal is to minimize the total time required to complete a single round of federated training. This metric, formally known as the makespan or round completion time, is a key indicator of system performance.

\subsubsection{Communication Resource Constraints}

The federated learning system in satellite network comprises a server $\mathbb{S}$, and a set of $|\mathcal{C} |$ satellite clients, denoted by the set $\mathcal{C} = \{\mathbb{C}_1, \mathbb{C}_2, \dots, \mathbb{C}_{|\mathcal{C}|}\}$. 
The federated learning process is iterative, proceeding in discrete rounds. Each global round is composed of three sequential phases: first, the distribution of the current global model; second, the local training of the model on the clients' private data; and third, the aggregation of the updated local models.
In the model distribution phase, the path $p_{\mathbb{S}, \mathbb{C}_i}$ represents the sequence of communication links that connect the server $\mathbb{S}$ to a client $\mathbb{C}_i$. Every individual link $(j, k)$ within the path $p_{\mathbb{S}, \mathbb{C}_i}$ is characterized by its available bandwidth, $b_{(j,k)}$. The effective end-to-end bandwidth of a path is limited by its most constrained segment, a concept known as the bottleneck link. Consequently, the effective bandwidth for the path $p_{\mathbb{S}, \mathbb{C}_i}$ is determined by the minimum bandwidth of all its constituent links. This is expressed formally as:
\begin{equation}
b_{p_{\mathbb{S}, \mathbb{C}_i}} = \min \left\{ b_{(j,k)} \mid (j,k) \in p_{\mathbb{S}, \mathbb{C}_i} \right\}.
\label{eq:downlink_bandwidth}
\end{equation}
This equation specifies that the effective bandwidth of the path from the server $\mathbb{S}$ to a client $\mathbb{C}_i$, denoted $b_{p_{\mathbb{S}, \mathbb{C}_i}}$, is the minimum value found among the bandwidths $b_{(j,k)}$ of all links $(j,k)$ that constitute the path $p_{\mathbb{S}, \mathbb{C}_i}$.

Based on this effective bandwidth $b_{p_{\mathbb{S}, \mathbb{C}_i}}$, we can calculate the time required for the server $\mathbb{S}$ to transmit the global model to client $\mathbb{C}_i$. This duration, denoted as $\tau_{\mathbb{S}, \mathbb{C}_i}$, is directly proportional to the size of the model data and inversely proportional to the path's effective bandwidth. Let $d_{\mathbb{C}_i}$ represent the data size of the model parameters (e.g., in megabits) destined for client $\mathbb{C}_i$. The transmission time $\tau_{\mathbb{S}, \mathbb{C}_i}$ is then given by:
\begin{equation}
\tau_{\mathbb{S}, \mathbb{C}_i} = \frac{d_{\mathbb{C}_i}}{b_{p_{\mathbb{S}, \mathbb{C}_i}}}.
\label{eq:downlink_time}
\end{equation}
This equation calculates the model distribution time for client $\mathbb{C}_i$ as the ratio of the model data size to the effective bandwidth of the communication path from the server to the client.

A similar formulation applies to the model aggregation phase. During this phase, each client $\mathbb{C}_i$ transmits its locally updated model parameters back to the server $\mathbb{S}$ along a return path, which we denote as $p_{\mathbb{C}_i, \mathbb{S}}$. 
It is important to recognize that network routing can be asymmetric, meaning $p_{\mathbb{C}_i, \mathbb{S}}$ may not be identical to $p_{\mathbb{S}, \mathbb{C}_i}$. 
The effective bandwidth of the path $p_{\mathbb{S}, \mathbb{C}_i}$, denoted by $b_{p_{\mathbb{C}_i, \mathbb{S}}}$, is likewise determined by its bottleneck link:
\begin{equation}
b_{p_{\mathbb{C}_i, \mathbb{S}}} = \min \left\{ b_{(j,k)} \mid (j,k) \in p_{\mathbb{C}_i, \mathbb{S}} \right\}.
\label{eq:uplink_bandwidth}
\end{equation}

The time required for client $\mathbb{C}_i$ to complete this model aggregation, denoted $\tau_{\mathbb{C}_i, \mathbb{S}}$, is subsequently calculated as:
\begin{equation}
\tau_{\mathbb{C}_i, \mathbb{S}} = \frac{d_{\mathbb{C}_i}}{b_{p_{\mathbb{C}_i, \mathbb{S}}}}.
\label{eq:uplink_time}
\end{equation}
For simplicity, we assume the size of the model uploaded by client $\mathbb{C}_i$ is the same as the size of the model it downloaded. Therefore, the local model size of client $\mathbb{C}_i$ is expressed as
\begin{equation}
d_{\mathbb{C}_i} = d_{\mathbb{M}},
\label{eq:model_size}
\end{equation}
where $d_{\mathbb{M}}$ denotes the size of the AI model.

\subsubsection{Federated Learning-Related Constraints}


To construct the federated learning scheduling problem, we introduce a set of decision variables that represent the start and end times for the key operations at each client. For every client $\mathbb{C}_i \in \mathcal{C}$, these decision variables are defined as follows:
\begin{itemize}
    \item $t^s_{(\mathbb{S}, \mathbb{C}_i)}$ and $t^e_{(\mathbb{S}, \mathbb{C}_i)}$: The start and end times, respectively, for the model distribution phase of client $\mathbb{C}_i \in \mathcal{C}$.
    \item $t^s_{\mathbb{C}_i}$ and $t^e_{\mathbb{C}_i}$: The start and end times, respectively, for the local model training phase at the client $\mathbb{C}_i \in \mathcal{C}$.
    \item $t^s_{(\mathbb{C}_i, \mathbb{S})}$ and $t^e_{(\mathbb{C}_i, \mathbb{S})}$: The start and end times, respectively, for the model aggregation phase client $\mathbb{C}_i \in \mathcal{C}$. 
\end{itemize}

Our optimization model is designed to find the optimal values for these time-based decision variables. To ensure that any solution corresponds to a feasible and logically coherent schedule, we must introduce a series of constraints.

\paragraph{Task Duration Constraints} 

The first set of constraints defines the duration allocated for model distribution to each client. For each client $\mathbb{C}_i$, the schedule must ensure that the time interval between the start and end of the transmission precisely matches the calculated model distribution time $\tau_{\mathbb{S}, \mathbb{C}_i}$. This guarantees that each client receives the complete model. Formally, this is expressed as
\begin{equation}
t^e_{(\mathbb{S}, \mathbb{C}_i)} - t^s_{(\mathbb{S}, \mathbb{C}_i)} = \tau_{\mathbb{S}, \mathbb{C}_i}, \quad \forall \mathbb{C}_i \in \mathcal{C}.
\label{eq:constraint_dist_time}
\end{equation}

Similarly, for the local computation phase, each client $\mathbb{C}_i$ must be allocated a time interval for model training that exactly matches its specific local training duration $\tau_{\mathbb{C}_i}$. This ensures that the local computation can be completed without interruption. The constraint is therefore written as
\begin{equation}
t^e_{\mathbb{C}_i} - t^s_{\mathbb{C}_i} = \tau_{\mathbb{C}_i}, \quad \forall \mathbb{C}_i \in \mathcal{C}.
\label{eq:constraint_train_time}
\end{equation}

Finally, for the model aggregation phase, it is required that the time interval for uploading the local model from each client $\mathbb{C}_i$ to the server $\mathbb{S}$ matches exactly the required upload transmission time $\tau_{\mathbb{C}_i, \mathbb{S}}$. This ensures that aggregation proceeds smoothly and without timing conflicts. The formal constraint is given by
\begin{equation}
t^e_{(\mathbb{C}_i, \mathbb{S})} - t^s_{(\mathbb{C}_i, \mathbb{S})} = \tau_{\mathbb{C}_i, \mathbb{S}}, \quad \forall \mathbb{C}_i \in \mathcal{C}.
\label{eq:constraint_agg_time}
\end{equation}

\paragraph{Precedence and Ordering Constraints} 

The next set of constraints enforces the logical sequence of operations inherent to the federated learning process. Each client’s tasks must be executed in a specific, non-overlapping order. Specifically, a client cannot start local training until it has received the complete global model from the server. This precedence requirement is formally expressed as
\begin{equation}
t^e_{(\mathbb{S}, \mathbb{C}_i)} \leq t^s_{\mathbb{C}_i}, \quad \forall \mathbb{C}_i \in \mathcal{C}.
\label{eq:constraint_order_1}
\end{equation}

Similarly, it is required that the model aggregation phase for each client $\mathbb{C}_i$ begins only after its local training is completed. This ensures that the model uploaded to the server reflects the final updated version resulting from the completed local training. The corresponding constraint is given by
\begin{equation}
t^e_{\mathbb{C}_i} \leq t^s_{(\mathbb{C}_i, \mathbb{S})}, \quad \forall \mathbb{C}_i \in \mathcal{C}.
\label{eq:constraint_order_2}
\end{equation}

\paragraph{Objective Function}
A single round of federated learning is successfully concluded only when the server has received the updated models from all participating clients. The total duration of the round is therefore determined by the client that finishes its upload last. To accelerate the overall training process, it is imperative to minimize this maximum completion time. This is a classic makespan minimization objective. We formulate this objective as the optimization problem \textbf{P1}:
\begin{equation}
\begin{split}
\textbf{P1:} \quad \min \quad  & \max \left(  \left\{ t^e_{(\mathbb{C}_i, \mathbb{S})} | \forall \mathbb{C}_i \in \mathcal{C} \right\} \right), \\
\text{s.t.} \quad & \eqref{eq:constraint_dist_time}-\eqref{eq:constraint_order_2}.
\end{split}
\label{eq:objective}
\end{equation}
The objective function in \textbf{P1} seeks to find a schedule that minimizes the makespan, which is defined as the maximum value among all model aggregation completion times $t^e_{(\mathbb{C}_i, \mathbb{S})}$ for every client $\mathbb{C}_i$ in the set $\mathcal{C}$.

The problem \textbf{P1} can thus be classified as a sophisticated scheduling problem subject to precedence constraints, sharing structural similarities with scheduling challenges found in large-scale data processing frameworks like MapReduce. However, the unique and defining challenge of our work stems from the dynamic satellite network environment. The problem is not merely one of scheduling tasks with predefined durations. It instead necessitates a holistic, joint optimization of communication resource allocation (which in turn determines the transmission task durations $\tau_{\mathbb{S}, \mathbb{C}_i}$ and $\tau_{\mathbb{C}_i, \mathbb{S}}$) and the subsequent scheduling of these transmission and local training tasks. The primary contribution of this work is to address this complex, integrated optimization problem within the dynamic, resource-scarce context of satellite networks.


\section{The Proposed Temporal Graph-Based On-Demand Scheduling Method} \label{sec:method}

This section presents the proposed on-demand scheduling scheme, which is specifically engineered for satellite-based federated learning environments. The proposed method is built upon a temporal graph framework that accurately models the time-varying nature of the satellite network. It seeks to minimize the average completion time of each learning round by jointly optimizing data transmission schedules with the local training timeline of the federated learning process. The architecture of our solution is composed of two integral components. The first is a temporal graph model designed to represent the dynamic satellite network topology. The second is a sophisticated scheduling algorithm that orchestrates model distribution, local training process, and model aggregation.


\subsection{Temporal Graph Modeling for Satellite Federated Learning}

To accurately capture the dynamic characteristics of the satellite network, we adopt a discrete-time representation \cite{zhang2023time, guo2024enhanced}. The entire operational period is partitioned into a sequence of discrete, equal-length time windows. Each time window, indexed by $\tau_x$, corresponds to a snapshot of the satellite network, which we model as a graph $\mathcal{G}_{\tau_x} = (\mathcal{V}_{\tau_x}, \mathcal{E}_{\tau_x})$. In this formulation, the vertex set $\mathcal{V}_{\tau_x}$ includes all satellites and  user devices at time $\tau_x$. The edge set $\mathcal{E}_{\tau_x}$ represents all available communication links, encompassing both inter-satellite links  and satellite-to-ground links during that specific time window.

Each edge $e \in \mathcal{E}_{\tau_x}$ is annotated with a tuple of its parameters, $(b_e, d_e)$, representing its available bandwidth and propagation delay, respectively. These attributes are inherently time-dependent, fluctuating across different time windows due to factors such as satellite orbital motion, atmospheric conditions, and changes in link availability. 
The collection of these sequential snapshot graphs constitutes a temporal graph, which we denote as ${G} = \{\mathcal{G}_{\tau_1}, \mathcal{G}_{\tau_2}, \dots, \mathcal{G}_{\tau_T}\}$. This explicit temporal representation is fundamental to our approach. It provides a structured model that enables the computation of time-dependent communication paths and their associated end-to-end metrics, such as achievable bandwidth and total propagation delay, between the server and any client at any time window.

\subsection{Scheduling Scheme for Satellite-based Federated Learning}

To accelerate each global round of federated learning, we introduce an on-demand scheduling scheme that fundamentally departs from the conventional statistical multiplexing paradigm. The core principle of our approach is to meticulously coordinate the exchange of model parameters by considering both the available bandwidth of satellite network paths and the local training requirements of each client. This coordination strategy is designed to schedule transmissions serially, which eliminates resource contention and allows certain clients to begin their computationally intensive training tasks much earlier.

\subsubsection{The Core Idea of On-Demand Scheduling}

To highlight the benefits of our proposed on-demand scheduling method, we first analyze the limitations of conventional statistical multiplexing scheme to handle model parameter distribution and aggregation in federated learning. Note that the timing diagrams of the conventional and proposed schemes, shown earlier in Figs.~\ref{fig:old_method} and \ref{fig:new_method}, can be found in Section I for comparison.

In particular, in conventional federated learning implementations, a statistical multiplexing approach is often implicitly adopted for model distribution and model aggregation, as shown in Figure \ref{fig:old_method}. During the model distribution phase, the server attempts to transmit the global model to all participating clients concurrently \cite{cheng2025scheduling, siew2024fair}. Because the server's available bandwidth is limited and shared resource, this concurrent model transmission forces all clients to contend for bandwidth. This contention results in a fractional allocation of bandwidth to each client, which significantly prolongs the time required for any single client to receive the complete model \cite{shi2024satellite}. Consequently, the commencement of all local training processes is delayed. A similar contention issue arises again during the model aggregation phase. Upon completing their local training, multiple clients simultaneously attempt to upload their updated local models to the server, once again creating a network bottleneck at the server's receiver and slowing down the entire model aggregation process. The combined effect of resource competition during both model distribution and aggregation leads to a substantially elongated round completion time.

In contrast, our proposed on-demand scheduling scheme, depicted in Figure \ref{fig:new_method}, introduces a disciplined, sequential access model. During both the distribution and aggregation phases, network resources are exclusively allocated to a single client-server transmission at any given time. Other clients are required to wait until the ongoing transmission is complete. By scheduling model distributions serially, the first client in the sequence receives the full model at the maximum possible data rate, allowing it to commence its local training significantly earlier than would be possible under a concurrent scheme. This head start in local training naturally leads to an earlier completion of the local training task. As a result, this client can begin its model upload process sooner, potentially during a time when network resources are uncontested by other clients. This cascading effect, where early model reception facilitates early training completion and subsequent early model aggregation, effectively staggers the use of network resources and avoids the bottlenecks that plague the statistical multiplexing approach.

\subsubsection{Algorithm Description}

Algorithm~\ref{alg:scheduling} presents the proposed on-demand scheduling scheduling procedure for model distribution and aggregation in satellite federated learning systems. The objective is to assign contention-free transmission schedules for all client-server communications while minimizing the total time for each global round of federated learning.

\begin{algorithm}[t]
  \caption{On-Demand Scheduling Algorithm for Large-Scale LEO Satellite Network-based  Federated Learning}
  \label{alg:scheduling}
  \hspace*{0.02in} \textbf{Input:}
  Federated learning task $F = \{\mathbb{S}, \mathcal{C}, \mathbb{M}\}$, where $\mathbb{S}$ is the server, $\mathcal{C} = \{\mathbb{C}_1, \dots, \mathbb{C}_{|\mathcal{C}|}\}$ is the client set,  and $\mathbb{M}$ is the AI model. Temporal graph ${G} = \{\mathcal{G}_{\tau_1}, \mathcal{G}_{\tau_2}, \dots, \mathcal{G}_{\tau_T}\}$.\\
  \hspace*{0.02in} \textbf{Output:}
  A complete schedule scheme for all model distribution and aggregation transmissions in satellite networks.
  \begin{algorithmic}[1]
  \item Record the start time as $t_0$.
    \item \textbf{for} each client $\mathbb{C}_i \in \mathcal{C}$ \textbf{do}
    \item \ \ \ \ Compute the path $p_{\mathbb{S}, \mathbb{C}_i}$ from server $\mathbb{S}$ to client $\mathbb{C}_i$.
    \item \ \ \ \ Compute $b_{p_{\mathbb{S}, \mathbb{C}_i}} = \min \left\{ b_{(i,j)} \mid (i,j) \in p_{\mathbb{S}, \mathbb{C}_i} \right\}$.
    \item \ \ \ \ Compute the transmission time $\tau_{\mathbb{S}, \mathbb{C}_i} = \frac{d_{\mathbb{M}}}{b_{p_{\mathbb{S}, \mathbb{C}_i}}}$.
    \item \textbf{end for}
    \item Sort all clients in \textit{ascending order} of transmission time, forming an ordered list $\mathcal{C}'$.
    \item \textbf{for} each client $\mathbb{C}_i \in \mathcal{C}'$ \textbf{do}
    \item \ \ \ \ Transmit the global model along $p_{\mathbb{S}, \mathbb{C}_i}$.
    \item \ \ \ \ \textbf{if} transmission completes \textbf{then}
    \item \ \ \ \ \ \ \ \ Start local training at client $\mathbb{C}_i$.
    \item \ \ \ \ \textbf{end if}
\item \ \ \ \ \textbf{if} no other client is exchanging model with $\mathbb{S}$ \textbf{then}
    \item \ \ \ \ \ \ \ \ Compute the path $p_{\mathbb{C}_i, \mathbb{S}}$.
     \item \ \ \ \ \ \ \ \  Start uploading the local model $\mathbb{M}$ via $p_{\mathbb{C}_i, \mathbb{S}}$.
    \item \ \ \ \ \textbf{end if}
    \item \ \ \ \ Record the model aggregation completion time $t^e_{( \mathbb{C}_i,\mathbb{S})}$.
        \item \textbf{end for}
    \item \textbf{return} Total round time $\max(\left\{ t^e_{(\mathbb{C}_i, \mathbb{S})} | \forall \mathbb{C}_i \in \mathcal{C}  \right\}) - t_0$.
  \end{algorithmic}
\end{algorithm}

\textbf{Initialization.} The algorithm begins by recording the initial timestamp $t_0$ to later calculate the total time required for one federated learning  round. The system is given a federated learning task $F = \{\mathbb{S}, \mathcal{C}, \mathbb{M}\}$, where $\mathbb{S}$ denotes the server, $\mathcal{C} = \{\mathbb{C}_1, \dots, \mathbb{C}_{|\mathcal{C}|}\}$ is the set of participating clients, and $\mathbb{M}$ is the AI model. The network topology is represented as a temporal graph $G= \{\mathcal{G}_{\tau_1}, \dots, \mathcal{G}_{\tau_T}\}$, where $\mathcal{G}_{\tau}$ denotes the snapshot graph at time window $\tau$.

\textbf{Model Distribution Phase.} For each client $\mathbb{C}_i \in \mathcal{C}$, the algorithm first determines the end-to-end transmission path $p_{\mathbb{S}, \mathbb{C}_i}$ from the server to the client. This path is selected from the temporal graph to ensure feasibility in both connectivity and bandwidth. The bottleneck bandwidth along this path is then computed as $b_{p_{\mathbb{S}, \mathbb{C}_i}} = \min \{ b_{(i,j)} \mid (i,j) \in p_{\mathbb{S}, \mathbb{C}_i} \}$,
where $b_{(i,j)} $ denotes the link capacity between satellite nodes $i$ and $j$ at time window $\tau$. The corresponding transmission time required to deliver the global model to client $\mathbb{C}_i$ is calculated by
$\tau_{\mathbb{S}, \mathbb{C}_i} = \frac{d_{\mathbb{M}}}{b_{p_{\mathbb{S}, \mathbb{C}_i}}}$,where $d_{\mathbb{M}}$ is the size of the global model.

\textbf{Client Ordering.} After computing the transmission time for each client, the clients are sorted in ascending order of $\tau_{\mathbb{S}, \mathbb{C}_i}$ to form an ordered list $\mathcal{C}'$. This sorting helps prioritize clients with shorter model delivery times, improving scheduling efficiency. 

\textbf{Scheduling Transmission and Training.} The algorithm proceeds sequentially through the sorted client list $\mathcal{C}'$. For each client $\mathbb{C}_i$, the global model is transmitted along the precomputed path $p_{\mathbb{S}, \mathbb{C}_i}$. Once the transmission is complete, client $\mathbb{C}_i$ immediately starts local training.

\textbf{Model Aggregation Phase.} After local training is complete, client $\mathbb{C}_i$ uploads the trained model back to the server $\mathbb{S}$. Model aggregation follows a first-finish, first-upload order, meaning the client that completes local training earliest begins aggregation first.
To avoid contention, this model aggregation is initiated only if no other client is currently performing a local model upload. The path $p_{\mathbb{C}_i, \mathbb{S}}$ is computed, and model aggregation begins once the transmission starts. The timestamp for the completion of the aggregation is recorded as $t^e_{(\mathbb{C}_i, \mathbb{S})}$.

\textbf{Termination.} After all clients have completed their upload and the server has received all local models, the algorithm computes the total time required for the federated learning  round as
$\max_{\mathbb{C}_i \in \mathcal{C}} \left\{ t^e_{(\mathbb{C}_i, \mathbb{S})} \right\} - t_0$.
This value is returned as the round completion time.

\subsubsection{Complexity Analysis} 
The \textbf{Algorithm 1} computes paths and transmission times for all clients, with complexity $O(|\mathcal{C}|\cdot (|\mathcal{E}_{\tau}| + |\mathcal{V}_{\tau}|\log|\mathcal{V}_{\tau}|))$, where $|\mathcal{V}_{\tau}|$ and $|\mathcal{E}_{\tau}|$ are the numbers of nodes and edges in the snapshot graph at time $\tau$, and shortest paths are computed using Dijkstra's algorithm. Sorting clients adds $O(|\mathcal{C}|\log |\mathcal{C}|)$, and scheduling transmissions and aggregations requires $O(|\mathcal{C}|)$. Thus, the overall complexity is $O(|\mathcal{C}|\cdot (|\mathcal{E}_{\tau}| + |\mathcal{V}_{\tau}|\log|\mathcal{V}_{\tau}| + \log |\mathcal{C}| +1))$, indicating the algorithm scales efficiently with the number of clients and graph size, suitable for large-scale LEO satellite networks.

\section{Performance evaluation} \label{sec:evaluation}

\subsection{Simulation Setup}  
To evaluate the effectiveness of the proposed temporal-graph-based federated learning scheduling scheme, we simulate a large-scale LEO satellite network based on the Starlink constellation setting. The network consists of 1,000 Starlink satellites forming a dynamic topology through inter-satellite links and satellite–user links. User terminals are placed in 12 geographically diverse cities, each representing a distinct region and network condition. Specifically, user devices are deployed in 12 different cities. Each city hosts one federated learning client, while a server is located in Singapore.
Link bandwidths are uniformly sampled from 10 to 30 Mbps. The federated learning  task is based on the CIFAR-10 dataset, which contains 60{,}000 color images of size 32$\times$32 pixels across 10 categories. The dataset is evenly partitioned into independent and identically distributed subsets, each assigned to a single client and retained locally throughout the training process to ensure data privacy.

We evaluate five representative AI models of varying sizes, with different transmission and training times. LeNet-5 has a model size of 0.3 MB and a local training time of around 25 seconds. MobileNetV2 \cite{sandler2018mobilenetv2} requires approximately 13.4 MB and 180 seconds, EfficientNet-B0 requires 20.3 MB and 300 seconds, ResNet-18 requires 44.7 MB and 480 seconds, and ResNet-34 requires 83.6 MB and 950 seconds. These models reflect a wide spectrum of real-world federated learning  workloads and enable a comprehensive assessment of scheduling performance under different computational and communication constraints. Unless otherwise specified, MobileNetV2 is used as the default AI model in the experiments.

\subsection{Simulation Results}

\begin{figure}[!t]
	\centering
	\includegraphics[width=66mm]{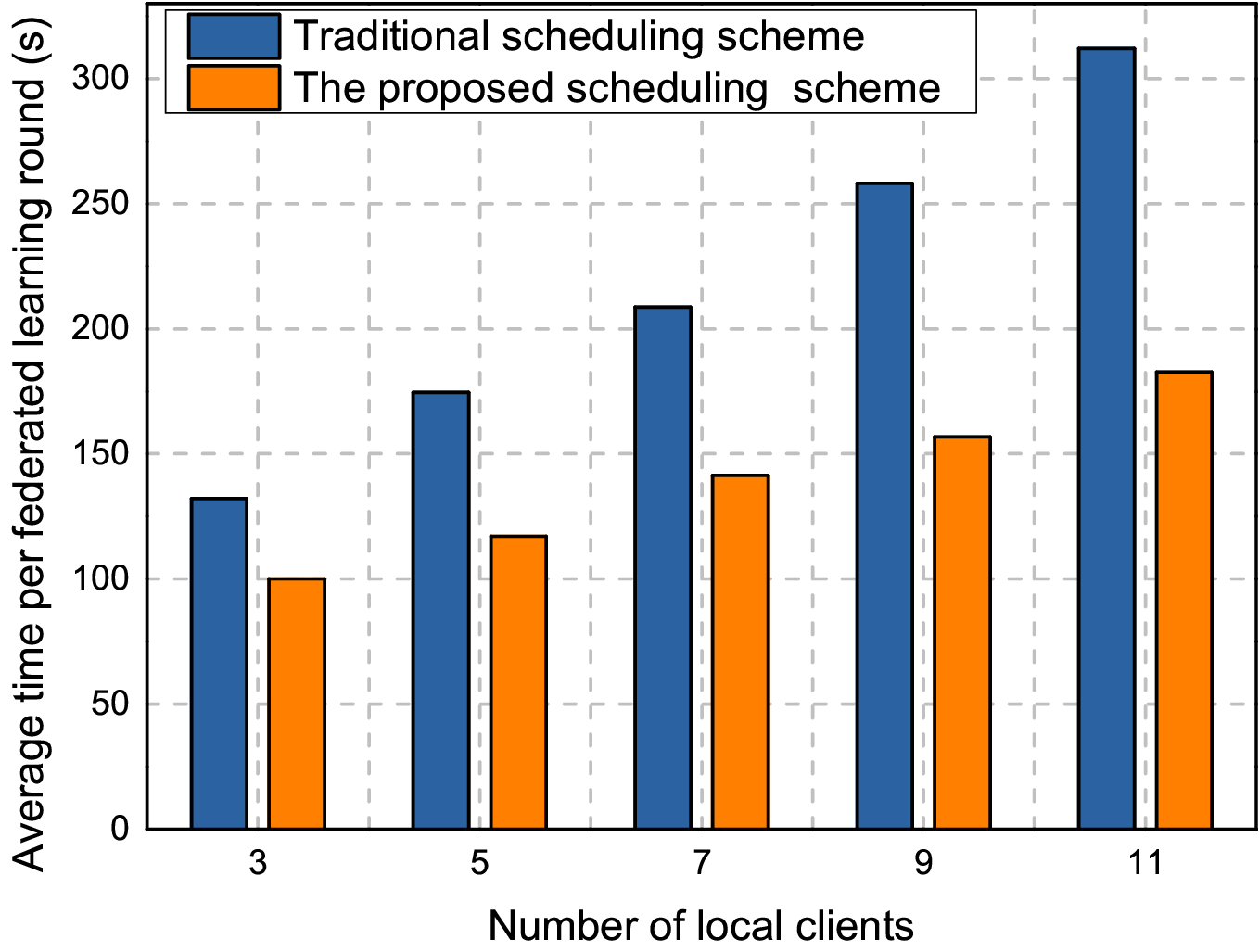}\\
	\caption{Average time per federated learning round under different  scheduling methods with local clients having identical computational capabilities.}
	\label{fig:times_versus_model_sizes}
\end{figure}

Firstly, by fixing the AI model to MobileNetV2, we evaluate the impact of different scheduling methods as the number of federated learning  clients increases. Fig.~\ref{fig:times_versus_model_sizes} presents the average time required for one federated learning  round under each scheduling strategy, assuming all clients have identical computational capabilities. The round time is measured from the moment the global model is dispatched to the completion of model aggregation after local training.
As the number of local clients increases, the average round time rises for both the traditional and the proposed on-demand scheduling methods. 
However, the proposed on-demand scheduling method consistently achieves lower round times, with reductions ranging from 24.27\% to 41.48\% compared to traditional methods.
This improvement is attributed to its ability to alleviate bandwidth contention during both model distribution and aggregation, which enables earlier initiation of local training and reduced overall duration. The performance gain becomes more significant at larger client scales, where traditional methods suffer from increasingly severe link contention. These results demonstrate the scalability and efficiency of the proposed on-demand scheduling approach in accelerating satellite-based federated learning.

\begin{figure}[!t]
	\centering
	\includegraphics[width=66mm]{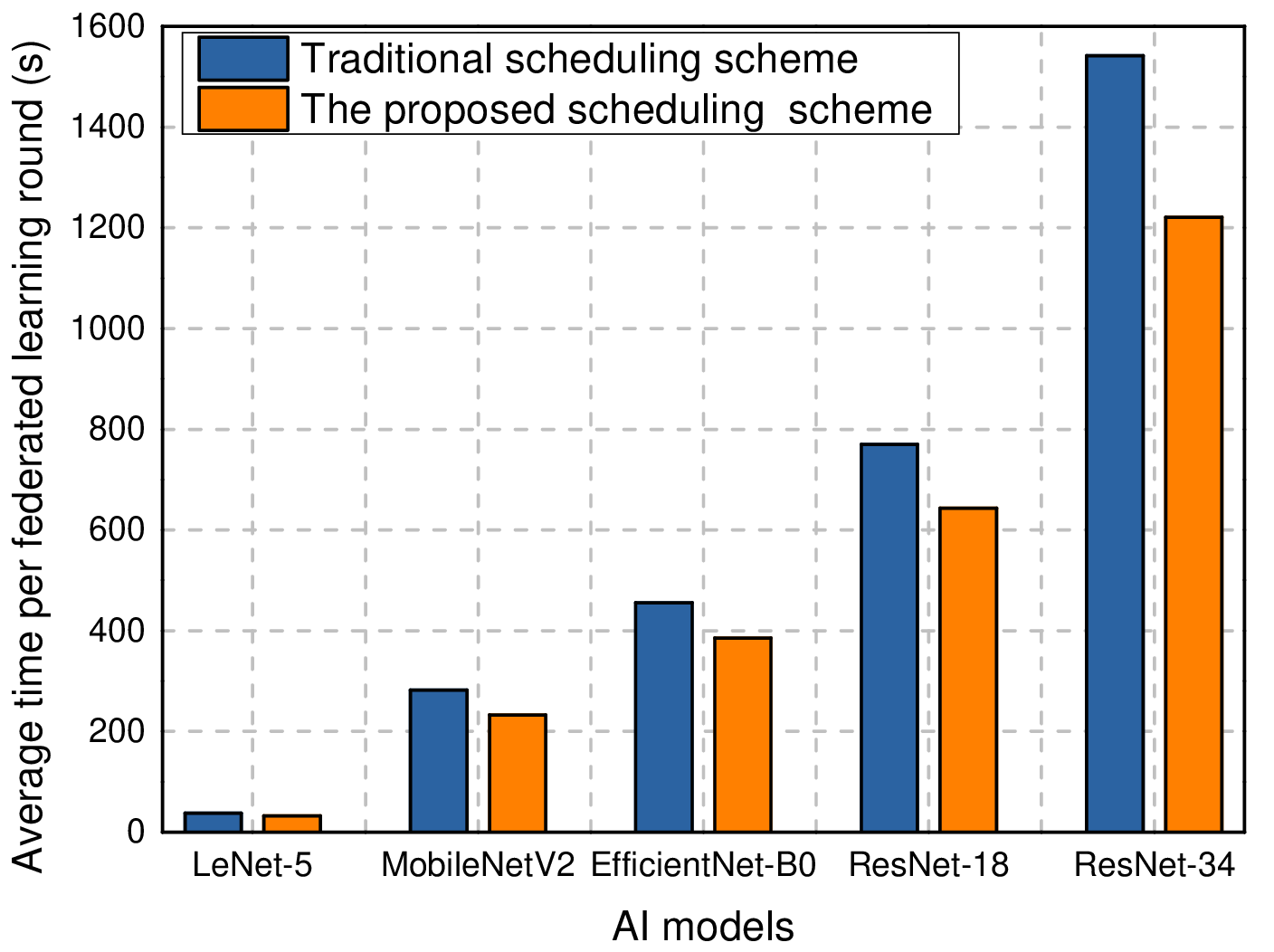}\\
	\caption{Average time per federated learning round under different scheduling methods versus different model sizes.}
	\label{fig:times_versus_client_numbers}
\end{figure}

Then, by fixing the number of local clients to five, we evaluate the impact of different AI model sizes on the performance of the proposed on-demand  scheduling scheme. Fig.~\ref{fig:times_versus_model_sizes} illustrates the average time required to complete one round of federated learning under the proposed on-demand  scheduling scheme and the traditional statistical multiplexing scheduling scheme, with five models arranged in ascending order of model size: LeNet-5 $<$ MobileNetV2 $<$ EfficientNet-B0 $<$ ResNet-18 $<$ ResNet-34. 
Across all AI models, the proposed on-demand scheduling scheme consistently achieves a shorter average round time compared to the traditional statistical multiplexing scheduling scheme, with reductions ranging from 14.20\% to 20.79\%.
Furthermore, the performance advantage becomes more pronounced as the AI model size increases. When the AI model is relatively small, the local training time is short, and the potential for optimization is limited. In contrast, for larger models with significantly longer training durations, the proposed on-demand  scheduling scheme is able to reduce a substantial portion of the communication overhead, leading to a greater reduction in the overall round time. These time savings accumulate over multiple training rounds, enabling a much faster completion of the entire federated learning process. Additionally, by reducing the time per round, the proposed on-demand  scheduling scheme allows more rounds to be completed within a given time window, which can result in improved model accuracy. These results demonstrate that the proposed on-demand  scheduling scheme is particularly beneficial for large-scale model training in satellite-based federated learning systems, and highlight its strong potential for enabling scalable and efficient AI model training across the world.

\vspace{-3 mm}

\section{Conclusion} \label{sec:conclusion}

This work investigated the scheduling of federated learning over large-scale LEO satellite networks, motivated by the growing need for AI model training across remote areas under privacy preserving and regulatory constraints. We analyzed the fundamental bottlenecks in model parameter distribution and aggregation caused by the dynamic topology and limited bandwidth of satellite networks. To address these issues, we proposed a discrete temporal graph–based on-demand scheduling framework that explicitly allocates bandwidth resources to reduce the makespan of each training round. 
Through extensive simulations, we demonstrated that the proposed method significantly outperformed the traditional statistical multiplexing scheme, with overall round times reduced by 14.20\% to 41.48\%. Even greater gains were observed for larger models and higher numbers of clients, thereby highlighting the scalability and practical value of the proposed on-demand scheduling scheme for accelerating federated learning in satellite networks.

\bibliographystyle{IEEEtran}
\bibliography{reference}

\end{document}